# AutoLugano: A Deep Learning Framework for Fully Automated Lymphoma Segmentation and Lugano Staging on FDG-PET/CT

Boyang Pan*[1,2], Zeyu Zhang*[3], Hongyu Meng*[4], Bin Cui[3], Yingying Zhang[3], Wenli Hou[3], Junhao Li[3], Langdi Zhong[5], Xiaoxiao Chen[5], Xiaoyu Xu[6], Changjin Zuo[3]#, Chao Cheng#[3], Nan-Jie Gong#[1,2]

[1]Institute of Magnetic Resonance and Molecular Imaging in Medicine, East China Normal University, Shanghai, China

[2]Shanghai Key Laboratory of Magnetic Resonance, School of Physics and Electronic Science, East China Normal University, Shanghai, China

[3] Department of Nuclear Medicine, Changhai Hospital, Naval Medical University, Shanghai 200433, China.

[4]Department of Radiology, Shanghai Fourth People's Hospital Affiliated to Tongji University, School of Medicine, Shanghai 200080, China.

[5]RadioDynamic Medical Shanghai 200062, China.

[6]Laboratory for Intelligent Medical Imaging, Tsinghua Cross-Strait Research Institute, Xiamen, China

#: these authors are corresponding authors. E-mails:13501925757@163.com; cjzuo@smmu.edu.cn; nanjie.gong@gmail.com

*: these authors contribute equally

## Abstract

**Purpose:** To develop a fully automated deep learning system, AutoLugano, for end-to-end lymphoma classification by performing lesion segmentation, anatomical localization, and automated Lugano staging from baseline FDG-PET/CT scans.

**Methods:** The AutoLugano system processes baseline FDG-PET/CT scans through three sequential modules: (1) Anatomy-Informed Lesion Segmentation, where a 3D nnU-Net model, trained on multi-channel inputs (CT, PET, and liver-normalized PET), performs automated lesion detection (2) Atlas-based Anatomical Localization, which leverages the TotalSegmentator toolkit to map segmented lesions to 21 predefined lymph node regions using deterministic anatomical rules; and (3) Automated Lugano Staging, where the spatial distribution of involved regions is translated into Lugano stages (I–IV) and therapeutic groups (Limited vs. Advanced Stage) according to the Lugano 2014 criteria. The system was trained on the public autoPET dataset (n=1,007) and externally validated on an independent cohort of 67 patients. Performance was assessed using accuracy, sensitivity, specificity, F1-score for regional involvement detection and staging agreement.

**Results:** On the external validation set, the proposed model demonstrated robust performance, achieving an overall accuracy of 88.31%, sensitivity of 74.47%, Specificity of 94.21% and an F1-score of 80.80% for regional involvement detection, outperforming baseline models. Most notably, for the critical clinical task of therapeutic stratification (Limited vs. Advanced Stage), the system achieved a high accuracy of 85.07%, with a specificity of 90.48% and a sensitivity of 82.61%. These results underscore the model's reliability in distinguishing early-stage from advanced-stage disease.

**Conclusion:** AutoLugano represents the first fully automated, end-to-end pipeline that translates a single baseline FDG-PET/CT scan into a complete Lugano stage. This study demonstrates its strong potential to assist in initial staging, treatment stratification, and supporting clinical decision-making.

## Introduction

Accurate staging of lymphoma is critical to the optimal management of the disease, as it directly dictates therapeutic strategies and predicts treatment response. The Lugano classification system was established to define lymphoma stages and guide subsequent treatment decisions based on positron emission tomography–computed tomography (PET-CT) information[1][2]. However, the current clinical practice of Lugano staging remains a labor-intensive and time-consuming process. It requires clinicians to meticulously identify and anatomically localize all involved lymph node regions across the entire body on FDG-PET/CT scans, a task that demands substantial effort from specialized physicians and introduces significant inter-observer variability, leading to challenges in achieving consistent and reproducible staging results[3-5].

Ideally, a fully automated staging system would replicate the clinical workflow through three integrated subsystems: lesion segmentation, anatomical localization, and clinical staging logic. Although the advent of deep learning has accelerated computer-aided diagnosis (CAD) for lymphoma, most existing systems have concentrated on isolated tasks such as lesion segmentation or the binary classification of individual lymph nodes*[6-13]*. While valuable, these approaches provide limited utility for comprehensive anatomical staging, which relies on a holistic assessment of the spatial distribution of involved nodes. Recent innovations, such as the longitudinal assessment system by Jemaa et al. *[14]* and the AI-assisted coPERCIST tool*[15]*, have attempted to bridge this gap by automating treatment response evaluation (e.g., quantifying metabolic changes). However, these systems are specifically designed for monitoring therapeutic outcomes and cannot derive the initial Lugano stage (I–IV)*[16]* from a single baseline scan. A pivotal bottleneck impeding the development of fully automated staging systems is the critical absence of publicly available, pixel-wise annotated datasets for the comprehensive set of lymph node regions mandated by Lugano. This lack of annotated data has rendered the preferred data-driven approach—training a dedicated model for precise lymph node region segmentation—currently infeasible*[17]*.

To overcome these limitations, we propose AutoLugano, a comprehensive deep learning framework for end-to-end lymphoma classification. This system represents the first fully automated pipeline capable of translating a single baseline FDG-PET/CT scan directly into a complete Lugano classification. The framework addresses the aforementioned challenges through three novel sequential modules designed to integrate lesion detection, anatomical mapping, and clinical reasoning into a cohesive workflow.

1.1 Anatomy-Informed Lesion Segmentation

Accurate lesion segmentation from FDG-PET/CT scans is essential for lymphoma staging, yet existing deep learning approaches often struggle to distinguish small, metabolically active lesions from areas of physiological uptake or artifacts*[18]*. This limitation is particularly pronounced in models that rely solely on PET intensity values, which are subject to inter-patient metabolic variability and may lead to false positives in regions such as bowel or urinary system. While multi-modal models incorporating CT have improved anatomical localization, they frequently lack mechanisms to explicitly incorporate clinical standardization practices—such as the liver-based reference recommended by the Lugano classification—thereby limiting their consistency and clinical alignment.

To address these issues, we developed an anatomy-informed segmentation model based on the state-of-the-art nnU-Net architecture *[19]*. Our approach integrates structural prior knowledge through a multi-channel input strategy comprising the original PET, CT, and a liver-normalized PET image. The liver normalization, automatically derived using the TotalSegmentator [20] toolkit, aligns with Lugano/Deauville standardization, reducing inter-subject metabolic variability and enhancing lesion conspicuity. Simultaneously, anatomical constraints from CT are utilized to suppress false positives in physiologically confounding or anatomically implausible regions.

1.2 Atlas-Based Anatomical Localization

To address the challenge of anatomical mapping without pixel-level lymph node annotations, the second module features a pragmatic, Atlas-based anatomical localization system. This module leverages the TotalSegmentator to automatically segment major organs and musculoskeletal structures from the CT scan. By utilizing these structures as anatomical landmarks, we constructed a deterministic mapping system that assigns detected lesions to 21 predefined lymph node regions[21] based on expert-defined spatial relationships. This innovation effectively circumvents the need for a prohibitively resource-intensive annotated dataset, enabling precise anatomical localization through a transparent and computationally efficient approach.

1.3 Automated Lugano Staging

The final module automates the clinical reasoning process by translating the spatial distribution of involved regions into a Lugano stage. By doing so, this module fills a significant gap, as no prior automated systems have incorporated the complete clinical logic of the Lugano criteria. Following the Lugano 2014 criteria [2], the system analyzes the involvement of regions relative to the diaphragm (e.g., unilateral vs. bilateral) and the presence of extranodal disease to automatically assign a final clinical stage (I–IV). Additionally, it stratifies patients into therapeutic groups—classifying Stages I and II as Limited Stage and Stages III and IV as Advanced Stage[16]—providing a comprehensive, clinically actionable output that supports initial treatment stratification.

# Method

## 2.1 Data Collection

### 2.1.1 Lesion Segmentation Training Dataset

For the development and training of our primary lesion segmentation model, we utilized a large-scale, publicly available dataset comprising 1,007 anonymized 18F-FDG-PET/CT scans from the 2023 autoPET grand challenge*[22]*. This dataset, which includes a diverse range of tumor cases with corresponding expert-delineated lesion masks, provided a robust foundation for training a generalizable segmentation algorithm.

**2.1.2 External Validation Dataset**

An independent, external validation cohort of 67 patients was retrospectively enrolled from Changhai Hospital between January 2021 and December 2023. Inclusion criteria for this cohort were: (1) age ≥ 18 years; (2) a diagnosis of lymphoma confirmed by histopathological analysis of a biopsy specimen; and (3) evidence of FDG-avid disease (i.e., high metabolic activity) on the PET/CT scan; . Patients were excluded if their scans exhibited severe motion artifacts or metallic implants that degraded image quality. The study protocol was approved by the Institutional Review Board of Changhai Hospital (Approval No. CHEC2025-038), and the requirement for informed consent was waived due to the retrospective nature of the analysis and the use of fully de-identified data.

Table1. Total number of patients diagnosed clinically in each Lugano staging.

| **Lugano staging** | I | II | III | IV |
|---|---|---|---|---|
| The number of patients | 8 | 13 | 17 | 29 |
| Lugano response | Limited Stage (I/II) | | Advanced Stage (III/IV) | |
| The number of patients | 21 | | 46 | |

**2.1.3 FDG-PET/CT Acquisition and Annotation**

All external validation scans were performed on a Biograph mCT scanner (Siemens Healthineers, Erlangen, Germany). Patients were instructed to fast for at least 6 hours before receiving an intravenous injection of 18F-FDG at a dose of 3.7 MBq/kg. Imaging was performed approximately 60 minutes post-injection. CT data were acquired using a low-dose protocol (120 kVp, 30 mAs) and PET data were acquired with an emission time of 2 minutes per bed position. Images were reconstructed using an ordered subset expectation maximization (OSEM) algorithm with time-of-flight (TOF) and point-spread function (PSF) correction.

For the establishment of the reference standard, all 67 validation scans were independently reviewed by two board-certified nuclear medicine physicians, each with over 10 years of experience in oncologic PET/CT interpretation. For each scan, the physicians annotated the presence or absence of lymphomatous involvement within each of the 21 pre-defined lymph node regions. Based on this regional assessment, a final Lugano stage (I-IV) was assigned. In cases of disagreement, a third senior physician with 20 years of experience provided a final consensus judgment. The distribution of patients according to clinical Lugano staging is presented in Table 1.

## 2.2 Overview of AutoLugano System

The proposed AutoLugano system is a fully automated pipeline designed to process a baseline FDG-PET/CT scan and output a complete Lugano stage. This integrated framework directly addresses a critical bottleneck in clinical practice by automating the traditionally labor-intensive and subjective process of lymphoma staging. The overview of the Workflow of the AutoLugano system is presented in Figure 1.

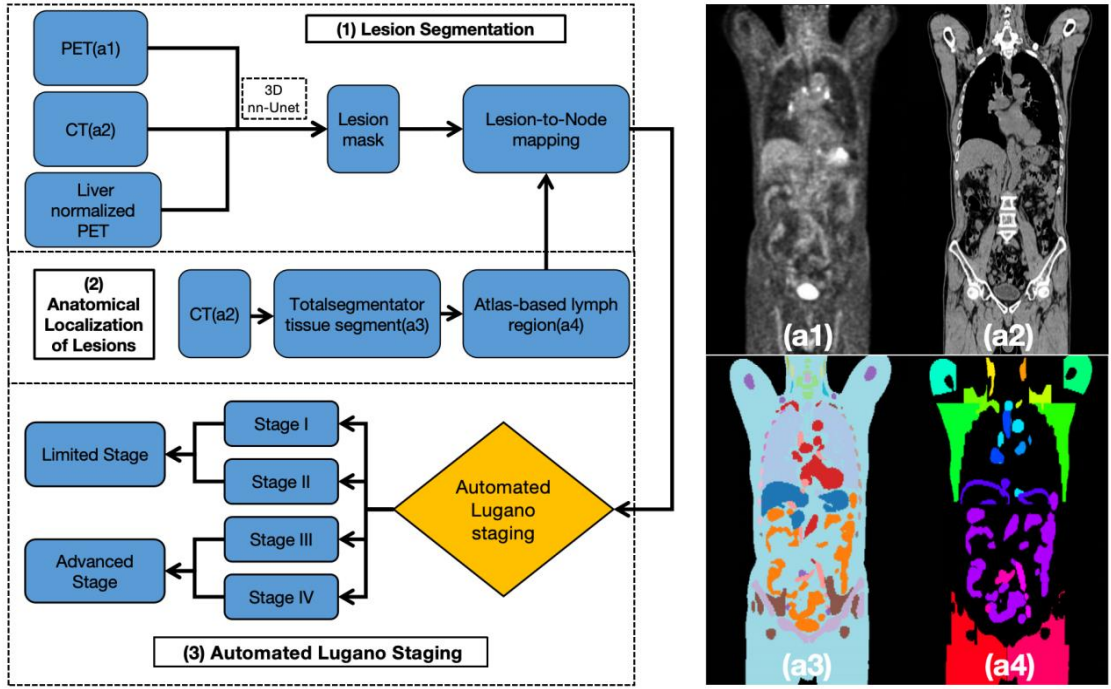

Figure 1. Workflow of the AutoLugano system, The processing procedure consists of three sequential modules: (1) **Lesion Segmentation -** where a 3D nnU-Net model, augmented with structural prior knowledge, automatically segments lymphoma lesions from the multi-channel input of PET (a1), CT (a2), and liver normalized PET; (2) **Anatomical Localization -** segmented lesions are mapped to 21 predefined lymph node regions (a4) by leveraging the tissue masks (a3) and Atlas-based lymph region atlas; (3) **Automated Lugano Staging -** the spatial distribution of involved regions is analyzed according to Lugano 2014 criteria to automatically assign the final disease stage (I – IV) and therapeutic group (Limited vs. Advanced Stage).

### 2.2.1 Lesion Segmentation with Structural Prior

The core of our pipeline is a lesion segmentation model based on the nnU-Net framework, which has demonstrated state-of-the-art performance across a wide range of medical imaging tasks. To enhance performance, we introduced a structural prior. First, for preprocessing, the TotalSegmentator tool was used to automatically segment the liver on the CT image. The mean and standard deviation of SUV within the liver were calculated and used to normalize the entire PET volume.

The 3D full-resolution nnU-Net was then trained on the autoPET dataset using a multi-channel input: (1) the original PET image, (2) the resampled CT image, and (3) the liver-normalized SUVnorm image. Standard nnU-Net [23] data augmentation schemes, including random rotations, scaling, and elastic deformations, were employed.

The model was optimized using a composite loss function combining Dice loss and Focal loss to handle the significant class imbalance between small lesions and the background. The final segmentation mask was generated by applying a probability threshold of 0.5 to the model's softmax output.

**2.2.2 Anatomical Localization of Lesions**

Accurately mapping the segmented lymphoma lesions to their specific anatomical lymph node regions is paramount for correct Lugano staging. We developed a Atlas-based anatomical localization module. This module leverages a comprehensive anatomical atlas automatically generated by the TotalSegmentator framework and its specialized extensions for fine-grained structures[24][25]. This process defines the 21 target lymph node regions based on their consistent spatial relationships with surrounding organs, vasculature, and bones. The specific regions were shown in table 3. For instance, the mediastinal region is defined as the space bounded by the segmented aorta, pulmonary artery, trachea, and heart. In contrast, regions like the popliteal fossa, which lack distinct organ boundaries, are defined using musculoskeletal landmarks, specifically the space posterior to the distal femur and proximal tibia.

All 21 region definitions were developed in consultation with our clinical experts and validated to ensure they are anatomically accurate and mutually exclusive[26]. Once the lesion masks and the 21 regional masks are generated, each lesion is assigned to a primary lymph node region based on maximal volumetric overlap. Furthermore, if a lesion's volume overlaps with a secondary or tertiary region by more than 25% of its overlap with the primary region, those regions are also considered involved.

**2.2.3 Automated Lugano Staging**

The final module of the AutoLugano framework automates the assignment of the Lugano stage by translating the spatial distribution of involved lymph node regions，as determined in the previous step, into a clinical stage according to the Lugano 2014 classification criteria [2]. The staging logic is implemented as follows:

Stage I: Involvement of a single lymph node region.

Stage II: Involvement of two or more lymph node regions on the same side of the diaphragm.

Stage III: Involvement of lymph node regions on both sides of the diaphragm.

Stage IV: Diffuse or disseminated involvement of one or more extranodal organs, with or without associated nodal involvement.

The spatial grouping of the 21 predefined lymph node regions relative to the diaphragm is fundamental to this logic. The regions superior to the diaphragm are considered as one side, and include: Waldeyer's ring, left and right neck, left and right clavicular areas, left and right axillae (armpits), left and right epitrochlear areas, the mediastinum, and the left and right pulmonary hila. The regions inferior to the diaphragmconstitute the other side, and include: the spleen, upper abdomen, lower abdomen, left and right para-iliac areas, left and right groins, and the left and right popliteal fossae.

For the purpose of identifying Stage IV disease, extranodal organ involvement is defined as focal or diffuse pathological FDG uptake within parenchymal organs (e.g., liver, lung) or the skeletal system.

Finally, the system categorizes the output stage into a therapeutic group: Stages I and II are classified as Limited Stage, while Stages III and IV are classified as Advanced Stage.

### 2.3 Statistical Analysis

The performance of the AutoLugano system was evaluated hierarchically across three critical endpoints corresponding to its core modules: (1) regional involvement detection accuracy, assessing the system's ability to correctly identify diseased lymph node regions; (2) final Lugano staging concordance, measuring the agreement between the automated pipeline and the physician consensus reference standard; and (3) therapeutic stratification performance, evaluating the system's utility in distinguishing Limited Stage (I/II) from Advanced Stage (III/IV) disease. All statistical analyses were performed using Python with the SciPy and scikit-learn libraries. In addition to conventional statistical tests, model performance was comprehensively evaluated using the F1-score, and the results are reported with 95% confidence intervals. A p-value of less than 0.05 was pre-specified as the threshold for statistical significance.

### 2.3.1 Regional Involvement Detection Performance

The foundational performance of the algorithms was evaluated based on their ability to correctly classify each of the 21 predefined lymph node regions as involved or not involved across the validation cohort. To contextualize our model's performance, we compared our proposed nnU-Net with a structural prior against two established baseline architectures: a default nnU-Net [20][27] with PET only input and PET/CT multiple input.

On a per-patient basis, segmentation performance was evaluated using Accuracy, Recall and F1-score to evaluate the rates of false positives and false negatives, respectively. The formula were as follows:

$$Accuracy = \frac{TP}{TP + FP} \tag{1}$$

,where TP denotes True Positive and FP denotes False Positive.

$$Recall = \frac{TP}{TP + FN} \tag{2}$$

,where FN denotes False Negative.

$$F1 - score = \frac{2TP}{2TP + FP + FN} \tag{3}$$

### 2.3.2 Automated Lugano Staging Performance

The accuracy of the system in replicating the final clinical stage was the primary clinical endpoint. The concordance between the automated Lugano stage (I, II, III, or IV) and the physician consensus reference standard was evaluated using several complementary metrics. First, a 4x4 confusion matrix was constructed to provide a detailed visualization of classification performance by comparing the automated stage predictions against the reference standard for each of the four stages. This analysis allows for the identification of specific error patterns, and from this matrix, we derived the per-class sensitivity (recall), specificity, and precision for each individual stage. Second, the overall staging accuracy was calculated as the percentage of patients for whom the automated stage was identical to the reference stage. Third, the quadratically weighted Kappa statistic was calculated to measure agreement while accounting for the ordinal nature of the staging system, which penalizes large disagreements more heavily than minor ones.

### 2.3.3 Therapeutic Response Stratification

The system's clinical utility in guiding treatment decisions was evaluated by assessing its ability to perform therapeutic risk stratification. This binary classification task, which categorizes patients into either Limited Stage (Lugano I/II) or Advanced Stage (Lugano III/IV), was evaluated using a standard set of diagnostic accuracy metrics. The primary metrics included overall accuracy, sensitivity, specificity, positive predictive value (PPV), and negative predictive value (NPV), which were to be reported with their 95% confidence intervals (CIs).[25]

## Results

### 3.1 Regional Involvement Detection Performance

The foundational performance of the automated algorithms was assessed by their ability to detect disease involvement within 1,407 individual nodal regions across the 67-patient cohort. Our proposed model demonstrated superior performance compared to the other tested variants, achieving the highest overall accuracy and an optimal F1-score by balancing sensitivity and specificity. The nnunet-pet-ct model also performed strongly, whereas threshold model showed a drop-off in performance. A summary comparison is presented in Table 2 and Figure 2.

Table 2: Overall performance comparison of different algorithms for regional disease detection.

| model | Accuracy(%)(95% CI) | Sensitivity (%)(95% CI) | Specificity (%)(95% CI) | false positive | false negative | F1-score(%)(95% CI) |
|---|---|---|---|---|---|---|
| AutoLugano | 88.31(86.52-90.10) | 74.47(72.14-76.80) | 94.21(92.93-95.49) | 58 | 109 | 80.80(78.76-82.84) |
| nnunet-pet-ct | 86.69(84.80-88.58) | 70.26(67.83-72.69) | 93.71(92.37-95.05) | 63 | 127 | 77.62(75.48-79.76) |
| nnunet-pet | 82.00(79.87-84.13) | 57.61(54.97-60.25) | 92.41(90.95-93.87) | 76 | 181 | 67.67(65.38-69.96) |
| threshold | 81.44(79.29-83.59) | 59.48(56.87-62.09) | 90.81(89.27-92.35) | 92 | 173 | 68.75(66.48-71.02) |

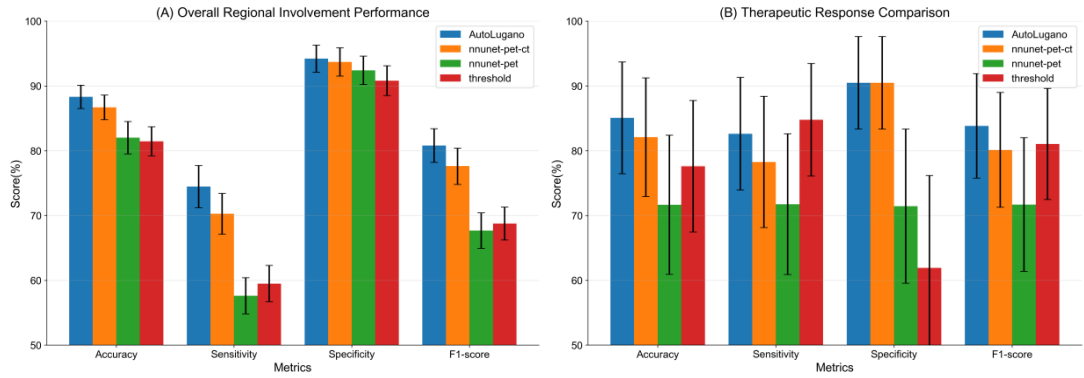

Figure 2. Performance of different models on overall regional involvement and therapeutic response. (A) illustrates the regional involvement detection performance (Accuracy, Sensitivity, Specificity, F1-score) of the proposed AutoLugano model against three baseline models (nnUNet-PET-CT, nnUNet-PET, and a threshold-based method). The proposed model shows the best performance across all metrics. (B) compares the therapeutic stratification (Limited vs. Advanced Stage) performance. The proposed AutoLugano achieves the highest accuracy (85.07%).

### 3.2 Per-Region Detection Performance

The performance of the proposed model varied considerably across the different anatomical regions (Table 3 and Figure 3). The model achieved excellent accuracy (>97%) in easily delineated regions such as the axilla and trochlea. In contrast, performance was notably lower in regions with complex anatomy and high physiological uptake, such as the Waldeyer's ring, where sensitivity was only 43.48%. The model achieved 100% accuracy in the popliteal fossa regions, though this was due to a lack of positive findings in the validation cohort. To visually demonstrate the model's performance across anatomically diverse lymph node regions, Figure 4 presents three representative clinical cases corresponding to key lymph node regions evaluated in this study.

Table 3: Per-region detection performance of proposed model.

| Anatomic Region | Accuracy(%)(95% CI) | Sensitivity(%)(95% CI) | Specificity(%)(95% CI) | false positive | false negative | positive case | F1-score(%)(95% CI) |
|---|---|---|---|---|---|---|---|
| Waldeyer's Ring | 80.88(71.52 – 90.24) | 43.48(23.78 – 63.18) | 100(91.48–100.00) | 0 | 13 | 24 | 56.56(39.12 – 74.00) |
| Left Neck | 83.82(73.58 – 94.06) | 75.00(61.38 – 88.62) | 96.43(87.66 – 100.00) | 1 | 10 | 40 | 79.17(67.34 – 91.00) |
| Right Neck | 79.41(69.11 – 89.71) | 69.23(56.48 – 82.00) | 93.10(84.91–100.00) | 2 | 12 | 40 | 73.97(62.34 – 85.60) |
| Left Infraclavicular | 77.94(67.58 – 88.30) | 60.00(45.93 – 74.07) | 85.42(75.68 – 95.16) | 7 | 8 | 20 | 67.80(54.34 – 81.26) |
| Right Infraclavicular | 85.29(75.48 – 95.10) | 66.67(53.33 – 80.01) | 88.68(79.34 – 98.02) | 6 | 5 | 15 | 74.84(63.34 – 86.34) |
| Left Axilla | 91.18(82.78 – 99.58) | 86.36(75.00 – 97.72) | 95.65(87.91–100.00) | 2 | 4 | 22 | 88.70(79.34 – 98.06) |
| Right Axilla | 97.06(90.94 – 100.00) | 94.74(85.38–100.00) | 97.96(90.15–100.00) | 1 | 1 | 19 | 95.89(88.70 – 100.00) |
| Left Trochlea | 98.53(94.17 – 100.00) | 100(100.00 – 100.00) | 98.51(91.01 – 100.00) | 1 | 0 | 1 | 99.26(95.00 – 100.00) |
| Right Trochlea | 98.53(94.17 – 100.00) | 100(100.00 – 100.00) | 98.46(90.94 – 100.00) | 1 | 0 | 3 | 99.26(95.00 – 100.00) |
| Mediastinum | 77.94(67.58 – 88.30) | 68.42(55.21–81.63) | 90(80.50 – 99.50) | 3 | 12 | 38 | 72.87(62.19-83.55) |
| Left Hilum | 91.18(82.78 – 99.58) | 68.75(53.13 – 84.37) | 98.08(90.15 – 100.00) | 1 | 5 | 16 | 78.39(68.71-88.07) |

| Region | | | | | | | |
|---|---|---|---|---|---|---|---|
| Right Hilum | 86.76(77.34 – 96.18) | 72.22(58.90 – 85.54) | 92.00(83.20 – 100.00) | 4 | 5 | 18 | 78.83(69.24-88.42) |
| Spleen | 85.29(75.48 – 95.10) | 66.67(53.33 – 80.01) | 90.57(81.34 – 99.66) | 5 | 5 | 15 | 75.53(65.22-85.84) |
| Upper Abdomen | 86.76(77.34 – 96.18) | 78.12(66.34 – 89.90) | 94.44(86.66 – 100.00) | 2 | 7 | 32 | 82.21(73.41-91.01) |
| Lower Abdomen | 83.82(73.58 – 94.06) | 78.95(66.34 – 91.56) | 90.00(80.50 – 99.50) | 3 | 8 | 38 | 81.31(72.31-90.31) |
| Left Para-iliac | 88.24(78.78 – 97.70) | 80.77(68.34 – 93.20) | 92.86(84.91 – 100.00) | 3 | 5 | 26 | 84.34(76.14-92.54) |
| Right Para-iliac | 85.29(75.48 – 95.10) | 73.08(60.00 – 86.16) | 92.86(84.91 – 100.00) | 3 | 7 | 26 | 78.71(69.09-88.33) |
| Left Groin | 85.29(75.48 – 95.10) | 83.33(70.00 – 96.66) | 86.00(76.34 – 95.66) | 7 | 3 | 18 | 84.30(76.09-82.51) |
| Right Groin | 80.88(71.52 – 90.24) | 61.11(47.78 – 74.44) | 88.00(78.34 – 97.66) | 6 | 7 | 18 | 69.62(58.68-80.56) |
| Left Popliteal Fossa | 100(94.63 – 100.00) | N/A | 100(94.63 – 100.00) | 0 | 0 | 0 | N/A |
| Right Popliteal Fossa | 100(94.63 – 100.00) | N/A | 100(94.63 – 100.00) | 0 | 0 | 0 | N/A |

* Sensitivity is marked N/A for regions with no positive cases in the validation set.

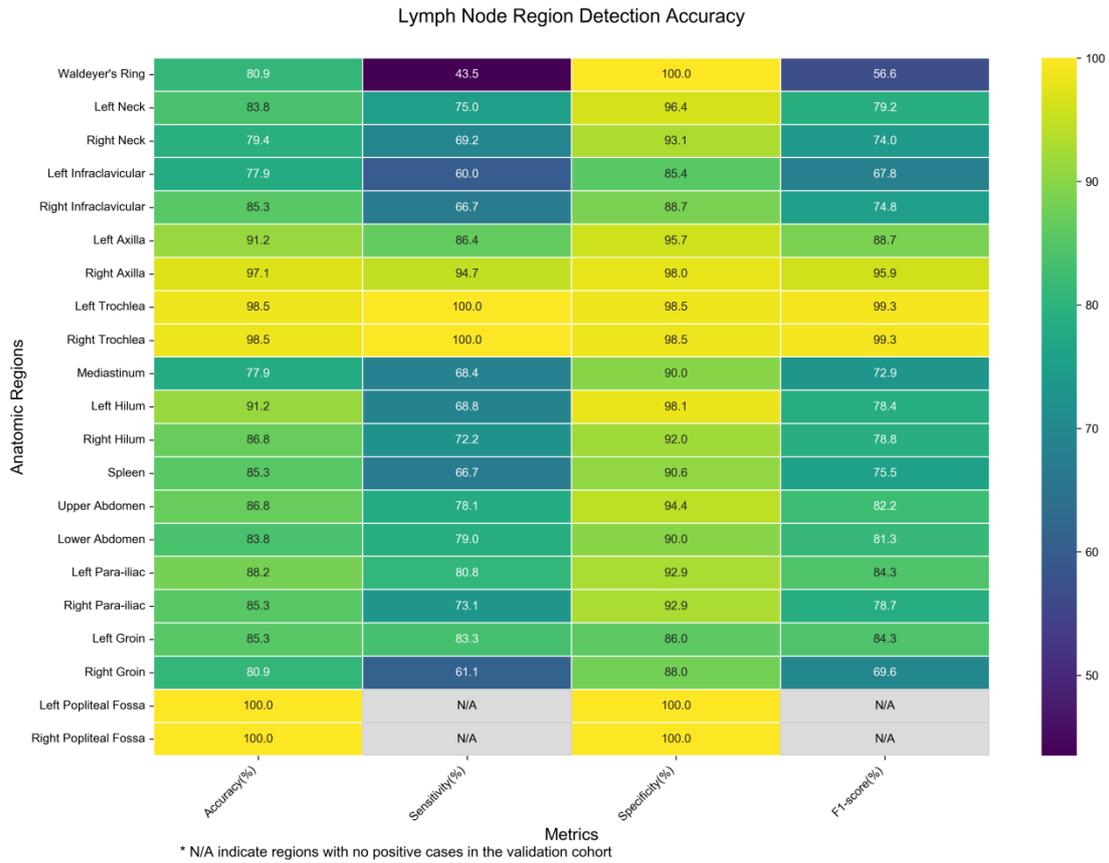

Figure 3. Heatmap of per-region detection sensitivity and accuracy for the proposed AutoLugano model. Rows represent the 21 predefined lymph node regions. Columns represent Accuracy (%), Sensitivity (%), Specificity (%) and F1-score(%). Notably, regions like Waldeyer's ring show lower sensitivity (43.48%) compared to regions like Right Axilla (>94%).

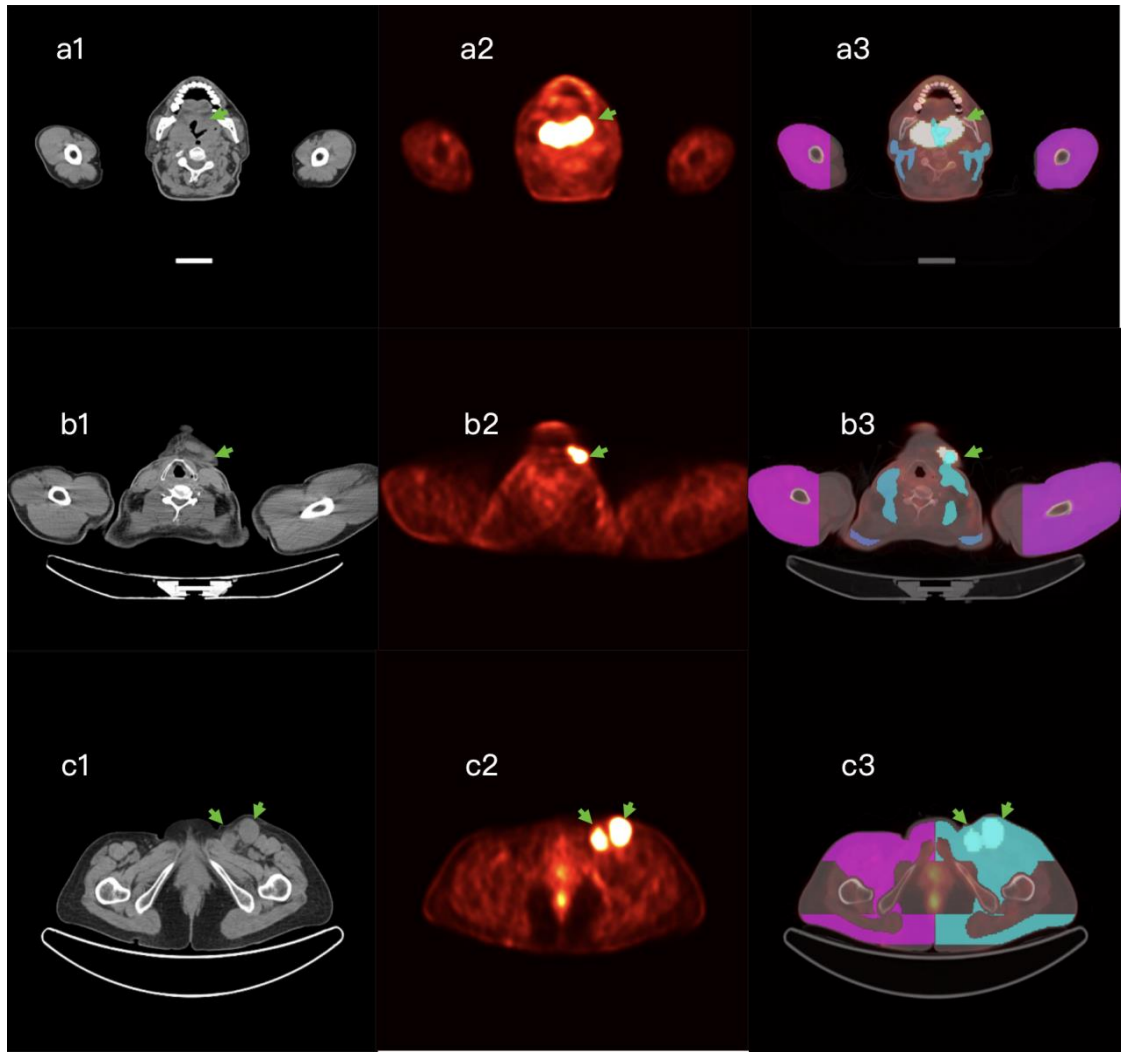

Figure 4. Automated lesion segmentation and anatomical localization using the AutoLugano system in representative lymphoma cases. The figure displays three cases: (a) a 65-year-old patient with lymphoma involving the Waldeyer's Ring; (b) a 47-year-old patient with lymphoma involving a left neck lymph node; and (c) a 59-year-old patient with lymphoma involving a left groin lymph node. For each case, axial CT in first columns (a1, b1, c1) providing anatomical context for the target lymph node; Corresponding FDG-PET in second columns (a2, b2, c2) demonstrating hypermetabolic activity of lesions. Automated lesion segmentation (red contours) in third columns (a3, b3, c3) overlaid on altas-based lymph node regions.

### 3.3 Automated Lugano Staging Performance

For the primary clinical endpoint of assigning the final Lugano stage, the proposed model achieved the highest overall accuracy of 62.69% (42 over 67 case). The agreement between the AutoLugano system and the physician reference standard was moderate (quadratically weighted Kappa = 0.49).

Figure 5 offers detailed confusion matrix into the performance of our AutoLugano staging. Overall , AutoLugano system demonstrates high accuracy. The diagonal cells, which indicate correct predictions, show consistently high values. The off-diagonal elements represent misclassifications. We can observe specific confusion patterns. Stage III and Stage IV are frequently confused with each other , indicating potential similarity between these two classes.

Table 4 details the performance of the AutoLugano system in the automated four-class Lugano staging task. Performance varied across the individual stages: it demonstrated higher sensitivity for Stage II (76.92%) and the highest accuracy for Stage IV (80.95%). However, the model showed limited in identifying Stage I and Stage III disease.

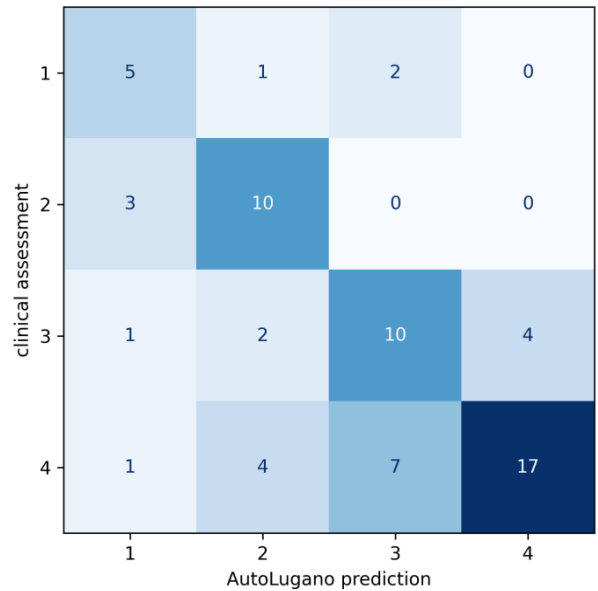

Figure 5. Confusion matrix illustrating the agreement between AutoLugano staging system and clinical assessment in four-class Lugano staging task.

Table 4: Comparison of overall accuracy for four-class Lugano staging.

| Lugano | Accuracy | Sensitivity | Specificity | F1-score(%)(95% |
|---|---|---|---|---|

| staging | (%)(95% CI) | (%)(95% CI) | (%)(95% CI) | CI) |
|---|---|---|---|---|
| I | 50.00(38.03-61.97) | 62.50(50.91-74.09) | 91.53(84.86-98.20) | 55.56(43.66-67.46) |
| II | 58.82(47.04-70.60) | 76.92(66.83-87.01) | 87.04(79.00-95.08) | 66.66(55.37-77.95) |
| III | 52.63(40.67-64.59) | 58.82(47.04-70.60) | 82.00(72.80-91.20) | 55.55(43.65-67.45) |
| IV | 80.95(71.55-90.35) | 58.62(46.83-70.41) | 89.47(82.12-96.82) | 68.00(56.83-79.17) |

## 3.4 Therapeutic Response Stratification Performance

For the critical task of stratifying patients into therapeutic response groups (Limited Stage vs. Advanced Stage), the prosed model and nnunet-PET-CT algorithms again demonstrated the best performance with an identical accuracy of 85.07%. These models were highly specific in identifying Limited Stage disease. Detailed performance metrics for the selected models are provided in Table 5 and Figure 2.

Table 5: Performance comparison for therapeutic response stratification.

| model | Accuracy (%)(95% CI) | Sensitivity (%)(95% CI) | Specificity (%)(95% CI) | PPV | NPV | F1-score(%)(95% CI) |
|---|---|---|---|---|---|---|
| AutoLugano | 85.07 (76.43–93.71) | 82.61 (73.91-91.31) | 90.48 (83.33-97.63) | 2 | 8 | 83.82 (75.76-91.88) |
| nnunet-pet-ct | 82.09 (72.93-91.25) | 78.26 (68.12-88.40) | 90.48 (83.33-97.63) | 2 | 10 | 80.13 (71.27-88.99) |
| nnunet-pet | 71.64 (60.90-82.38) | 71.74 (60.87-82.61) | 71.43 (59.52-83.34) | 6 | 13 | 71.69 (61.38-82.00) |

| threshold | 77.61 (67.46-87.76) | 84.78 (76.09-93.47) | 61.90 (47.62-76.18) | 8 | 7 | 81.04 (72.46-89.62) |

## Discussion

In this study, we developed and validated AutoLugano, the first fully automated, end-to-end deep learning framework capable of translating a single baseline FDG-PET/CT scan into a complete Lugano stage. By integrating lesion segmentation, anatomical localization, and clinical staging logic into a cohesive pipeline, our system achieved high accuracy in regional involvement detection (88.31%) and demonstrated potential for automated Lugano staging (accuracy 62.69%, weighted Kappa 0.49). Notably, for the critical task of therapeutic stratification (Limited vs. Advanced Stage), the model achieved a clinically relevant accuracy of 85.07%, underscoring its utility in informing initial treatment decisions. The system's performance and clinical relevance are best understood through its three core innovations: anatomy-informed lesion segmentation, Atlas-based anatomical localization, and automated staging logic.

To addresses the challenge of accurate lesion detection in the presence of physiological uptake, our enhanced nnU-Net model, which incorporates structural priors and multi-channel inputs (CT, PET, and liver-normalized PET), demonstrated superior performance compared to baseline architectures (PET-only and PET/CT inputs). This improvement highlights the critical value of integrating complementary anatomical and metabolic information. Specifically, the use of liver-normalized PET mitigates inter-patient variability in metabolic activity, thereby enhancing the consistency of lesion detection. Furthermore, the integration of structural priors significantly suppressed false positives in anatomically improbable regions, resulting in a high specificity of 94.21% for regional involvement—a crucial factor for preventing over-staging. The system also exhibited robustness in distinguishing pathological uptake from external artifacts, such as radiotracer contamination, which were correctly identified as extracorporeal through CT-based anatomical mapping.

Secondly, we successfully implemented a pragmatic solution to the bottleneck of anatomical localization, overcoming the scarcity of pixel-wise annotated lymph node datasets. Instead of relying on a fully supervised learning approach, which would require prohibitively expensive annotations, we developed a novel Atlas-based localization module. By leveraging the TotalSegmentator toolkit to segment major organs and musculoskeletal structures, we established a deterministic mapping system that encodes expert-defined spatial relationships. This approach is not only computationally efficient but also transparent. Analysis of per-region performance revealed excellent accuracy in anatomically distinct regions such as the axilla and trochlea. However, sensitivity was lower in complex areas like Waldeyer's ring, reflecting the inherent difficulty of segmenting lesions in regions with high physiological uptake or intricate anatomy. These findings suggest that while the Atlas-based approach is effective, future refinements should focus on these challenging anatomical sub-regions.

To our knowledge, this work introduces the first computational framework capable of automatically deriving a Lugano classification from the spatial distribution of involved lymph node regions. Unlike previous systems that focused primarily on lesion segmentation or longitudinal response assessment (e.g., Jemaa et al. *[14]* and coPERCIST*[15]*), AutoLugano is designed for baseline cross-sectional staging. It processes a single pre-treatment scan to determine the initial anatomical extent of the disease (Stage I–IV), which is the primary determinant for first-line therapeutic strategy. While the system demonstrated strong performance in binary therapeutic stratification, the error analysis revealed challenges in distinguishing Stage III from Stage IV disease. These misclassifications, which differentiate supradiaphragmatic lymphatic disease from disseminated extranodal involvement, likely stem from the sequential nature of the pipeline, where errors in segmentation or localization propagate to the final staging logic. Additionally, the model's limited explicit training for diverse extranodal disease patterns contributed to these discrepancies. Nevertheless, the system's ability to automate the complex clinical reasoning required for Lugano staging represents a significant advancement over existing CAD tools.

Interpretation of these results requires acknowledgment of the sequential nature of the pipeline, wherein errors in segmentation or localization propagate to the final staging output. Misclassification between Stage III and Stage IV, observed in our error analysis, carries significant clinical implications, as it differentiates supradiaphragmatic lymphatic disease from disseminated extranodal involvement. These errors likely arise from imprecise lesion localization relative to the diaphragm and limited explicit training for diverse extranodal disease patterns. Nevertheless, the model's strong performance in binary therapeutic stratification underscores its immediate clinical applicability for initial treatment selection.

While the results are promising, several limitations of this study must be acknowledged. First, the relatively small size of the external validation cohort (n=67) limits the generalizability of our findings, particularly for assessing performance in rare subtypes or underrepresented stages. Second, the dependency on the TotalSegmentator toolkit for anatomical mapping, while effective, introduces a potential point of failure if the tool segments key organs inaccurately. Third, as discussed, the current deterministic staging logic, while transparent and based on consensus criteria, is sensitive to segmentation inaccuracy. Fourth, the model was not explicitly trained to identify extranodal involvement beyond the defined lymphatic regions, which is a component of Stage IV disease. Finally, there exists a notable domain gap between the public autoPET dataset—which includes a variety of tumor types and segmentation targets—and the clinical lymphoma staging dataset used in our validation. This discrepancy may affect the model's generalization to real-world lymphoma staging scenarios.

Looking forward, AutoLugano holds significant potential for longitudinal analysis. By tracking lesions within precisely mapped anatomical regions over time, the framework could enable more sensitive, region-specific assessments of treatment response, offering granular insights into disease dynamics beyond whole-body metrics. With further refinement and validation on larger, multi-center cohorts, AutoLugano could evolve into an invaluable decision-support tool, enhancing the consistency and efficiency of lymphoma staging in clinical practice.

## Conclusion

AutoLugano presents a successful fully automated Lugano staging from a single FDG-PET/CT scan. It demonstrates compelling accuracy for therapeutic stratification and points out the challenges achieving precise anatomical staging. By automating this labor-intensive process, the framework holds significant promise as a decision-support technique to enhance consistency and efficiency in lymphoma staging, potentially aiding less experienced clinicians in resource-limited settings. With further refinement and validation, such systems could become invaluable components of the modern oncology workflow.